\newif\iffinal 
\finaltrue 

\newif\ifIR
\IRtrue

\iffinal
  \newcommand{\hans}[1]{}
  \newcommand{\maren}[1]{}
\else
  \newcommand{\hans}[1]{{\color{blue} \textbf{[Hans: #1]}}}
  \newcommand{\maren}[1]{{\color{orange} \textbf{[Maren: #1]}}}
\fi
\ifIR
  \newcommand{\hansIR}[1]{}
  \newcommand{\marenIR}[1]{}
\else
  \newcommand{\hansIR}[1]{{\color{blue} \textbf{[Hans: #1]}}}
  \newcommand{\marenIR}[1]{{\color{orange} \textbf{[Nico: #1]}}}
\fi

\documentclass{article}

\usepackage{microtype}
\usepackage{mathtools}
\usepackage{graphicx}
\usepackage{booktabs} 

\usepackage{xcolor}
\usepackage[caption=false]{subfig}
\usepackage{pgfplots}
\pgfplotsset{compat=1.12}
\usepgfplotslibrary{groupplots}

\definecolor{shadecolor}{RGB}{230,230,250}
\definecolor{plotblue}{RGB}{57, 106, 177}
\definecolor{plotorange}{RGB}{218, 124, 48}
\definecolor{plotgreen}{RGB}{62, 150, 81}
\definecolor{plotred}{RGB}{204, 37, 41}
\definecolor{plotgray}{RGB}{83, 81, 84}
\definecolor{plotpurple}{RGB}{107, 76, 154}
\definecolor{plotdarkred}{RGB}{146, 36, 40}
\definecolor{plotbeige}{RGB}{148, 139, 61}
\definecolor{plotnavy}{HTML}{001F3F}

\usepackage{tikz}
\usepgfplotslibrary{external}
\usepgfplotslibrary{statistics}
\usetikzlibrary{decorations}
\usetikzlibrary{decorations.markings}
\usetikzlibrary{pgfplots.statistics}
\pgfplotsset{compat = 1.3}

\pgfplotscreateplotcyclelist{mycolorlist}{%
plotblue,every mark/.append style={fill=plotblue, draw = plotblue!35!black, thin},mark=*, mark size=1.2pt\\%
plotorange,every mark/.append style={fill=plotorange, draw = plotorange!35!black, thin},mark=square*, mark size=1.2pt\\%
plotgreen,every mark/.append style={fill=plotgreen, draw = plotgreen!35!black, thin},mark=diamond*, mark size=1.6pt\\%
plotred, every mark/.append style={fill=plotred, draw = plotred!35!black, thin},mark=triangle*, mark size=1.2pt\\%
plotgray, densely dotted, every mark/.append style={fill=plotgray, draw = plotgray!35!black, thin},mark=*, mark size = 1.2pt\\%
plotpurple, densely dotted,every mark/.append style={fill=plotpurple, draw = plotpurple!35!black, thin},mark=square*\\%
plotdarkred, densely dotted,every mark/.append style={fill=plotdarkred, draw = plotdarkred!35!black, thin},mark=diamond*, mark size=1.2pt\\%
plotbeige, densely dotted,every mark/.append style={fill=plotbeige, draw = plotbeige!35!black, thin},mark=triangle*, mark size=1.2pt\\%
}
\pgfplotsset{cycle list name = mycolorlist}
\pgfplotsset{every axis plot/.append style={very thick}}
\pgfplotsset{grid = major, major grid style = {thin}}
\pgfplotsset{ticks = major, tick align=inside, tick style = {thin, gray}}
\pgfplotsset{compat=1.11}
\pgfplotsset{width=\columnwidth, height=0.6\columnwidth}
\pgfplotsset{legend style={at={(1,0)},xshift=-0.05cm, yshift=0.05cm, anchor=south east }}
\pgfplotsset{
  colormap={whitered}{color(0cm)=(plotorange!75!plotred); color(1cm)=(white)}
}
\usetikzlibrary{arrows.meta}

\usepackage{hyperref}

\usepackage{xr}
\usepackage{enumitem}

\usepackage{amsmath,amssymb,amsthm}
\usepackage{tensor}
\usepackage{xcolor}
\usepackage[nameinlink]{cleveref}
\usepackage[toc,page,titletoc]{appendix}
\usepackage{bbm}
\usepackage{mathtools}

\newcommand{\R}{\mathbb{R}}

\let\given\givenbase

\newcommand*{\defeq}{\coloneqq}

\newcommand*{\rd}{\mathrm{d}}

\newcommand*{\bs}{\boldsymbol}

\newcommand{\interior}[1]{%
  {\kern0pt#1}^{\mathrm{o}}%
}

\newlength{\figwidth}
\newlength{\figheight}

\crefname{supp}{Supplement}{Supplements}

\usepackage[accepted]{icml2020}

\icmltitlerunning{A Fourier State Space Model for Bayesian ODE Filters}

\begin{document}

\twocolumn[
\icmltitle{A Fourier State Space Model for Bayesian ODE Filters}




\icmlsetsymbol{intern}{*}

\begin{icmlauthorlist}
  \icmlauthor{Hans Kersting}{ekut,mpi,intern}
  \icmlauthor{Maren Mahsereci}{amazon}
\end{icmlauthorlist}

\icmlaffiliation{ekut}{University of T\"ubingen, Germany}
\icmlaffiliation{mpi}{Max Planck Institute for Intelligent Systems, T\"ubingen, Germany}
\icmlaffiliation{amazon}{Amazon Web Services, Berlin, Germany}

\icmlcorrespondingauthor{Hans Kersting}{hans.kersting@uni-tuebingen.de}

\icmlkeywords{Machine Learning, ICML}

\vskip 0.3in
]



\printAffiliationsAndNotice{$^*$work primarily performed during an internship at Amazon Research, Cambridge, UK.} 

\begin{abstract}
Gaussian ODE filtering is a probabilistic numerical method to solve ordinary differential equations (ODEs).
It computes a Bayesian posterior over the solution from evaluations of the vector field defining the ODE.
Its most popular version, which employs an integrated Brownian motion prior, uses Taylor expansions of the mean to extrapolate forward and has the same convergence rates as classical numerical methods.
As the solution of many important ODEs are periodic functions (oscillators), we raise the question whether Fourier expansions can also be brought to bear within the framework of Gaussian ODE filtering.
To this end, we construct a Fourier state space model for ODEs and a `hybrid' model that combines a Taylor (Brownian motion) and Fourier state space model.
We show by experiments how the hybrid model might become useful in cheaply predicting until the end of the time domain.
\end{abstract}

\section{Introduction}

Ordinary differential equations (ODEs) appear in many machine learning algorithms.
In recent years, there has been a particular surge of interest in ODEs for normalizing flows \citep{RezendeMohamed_2015}.
This development is driven by neural ODEs \citep{ChenDuvenaud18}, which allow for maximum-likelihood estimation and variational inference. 
Neural ODEs replace learning by gradient descent with learning by ODE sensitivity analysis \citep{Rackauckas2018SensitivityAnalysisInML}. \\
A recent recast of ODEs as a stochastic filtering problems has made it possible to solve initial value problems (IVPs) by all available Bayesian filtering methods \citep{TronarpKSH2019,TronarpSarkkaHennig_BayesianODESolvers_2020}.
The resulting class of methods, called \emph{ODE filters}, has not only fulfilled the goal of probabilistic numerics (PN) \citep{HenOsbGirRSPA2015,OatesSullivan2019} to quantify numerical uncertainty in a Bayesian way, but has also identified the dynamic model as the fundamental internal modeling assumption of ODE solvers.
This dynamic model determines how the solver extrapolates forward in time and is equivalent to a prior over the ODE solution \citep[Appendix A]{KerstingKraemer_godef_inverse_2020}.
\\ 
Early PN research has, to show that its new methods are indeed practical, focused mostly on creating probabilistic analogues of classical methods.
This line of inquiry has discovered that the integrated Brownian Motion (IBM) prior gives rise to ODE filters whose mean coincides with standard classical methods; see \citet{schober2019}. 
This is due to the fact that---like e.g.~Runge--Kutta method---ODE filters with the IBM prior use Taylor expansions to locally predict forward; see Equation (6) in \citet{KerstingSullivanHennig2018}.
In the meantime, other local expansions based on the Mat\'{e}rn covariance function have been studied; see \citet{TronarpSarkkaHennig_BayesianODESolvers_2020}.
Fourier expansions, however, have not been investigated in the context of ODE filters although it is known how to incorporate them in a dynamic model; see \citep{SolinSarkka_Periodic_2014}.
With this paper, we aim to begin filling in this gap.
Since so many important ODEs are oscillators with periodic solutions, we consider Fourier expansions a promising research direction in the context of ODE filtering.

\section{ODE Filtering for Initial Value Problems}

We consider the following IVP
\begin{align} \label{eq:ode}
  \dot{x}(t)
  =
  f \left( x(t) \right),\; \forall t\in [0,T], \qquad x(0)=x_0 \ \in \R^d,
\end{align}
with vector field $f:\R^d \to \R^d$.
For notational convenience, we restrict w.l.o.g.~the below presentation to $d=1$.
As the solution $x:[0,T] \to \R^d$ in general lies in an infinite dimensional function space such as $C^2([0,T];\R^d)$, a finite dimensional representation is needed to extrapolate from $x(t)$ to $x(t+h)$.
Runge--Kutta methods, for example, use Taylor expansions (i.e.~projections to polynomial spaces) as finite dimensional approximations of $x$.
While classical methods only do so implicitly, Gaussian ODE filtering  represents $x(t)$ explicitly in a $D$-dimensional \emph{state vector}, i.e.~in a stochastic process $X(t)$ from which a model of $x(t)$ can be linearly extracted:
\begin{align}
    x(t)
    \ \sim \
    H_0 X(t),
    \quad
    \text{ for some }
    H_0 \in \R^{d \times D}.
\end{align}
Moreover, for ODEs, the derivative has also to be linearly extractable
\begin{align}
    \dot{x}(t)
    \ \sim \
    H X(T)
    \quad
    \text{ for some }
    H \in \R^{d \times D}.
\end{align}
As ODE filtering is a Bayesian method, the state $X(t)$ is modeled by a stochastic process, which is usually represented by a linear time-invariant stochastic differential equation (SDE)
\begin{align}
    \label{eq:SDE}
    \rd X(t)
    =
    F X(t) \ \rd t
    +
    L \ \rd B(t)
\end{align}
with Gaussian initial condition on $X(0)$, where the drift and diffusion matrices $F,L \in \R^{D\times D}$ detail the deterministic and stochastic part of the dynamics respectively. 
This SDE prior can be thought of as a localized definition of a Gauss--Markov process. 
In its discretized form, it defines a \emph{dynamic model} 
\begin{align}  \label{eq:dynamic_model}
    p \left ( X(t+h) \given X(t)  \right )
    =
    \mathcal{N}(A(h)X(t), Q(h)),
\end{align}
with matrices $A(h),Q(h) \in \R^{D\times D}$ which are implied in closed form by $F$ and $L$.
To update this model, a \emph{measurement model} is added
\begin{align}  \label{eq:intractable_measurement_model}
    p(Z(t) \given X(t))
    &=
    \mathcal{N}\left(f(H_0 X(t)) - H X(t), R \right ),
\end{align}
$R \geq 0$, which is conditioned on the data
\begin{align}
    Z(t) 
    &\defeq 
    0.
\end{align}
As $f$ is non-linear, the above measurement model is intractable \citep[Section 2]{TronarpKSH2019}.
By substituting $f(H_0\mathbb{E}[X(t)])$ for $f(H_0 X(t))$, we obtain the following tractable measurement model
\begin{align}
    \label{eq:measurement_model}
    p(Z(t) \given X(t))
    &=
    \mathcal{N}\left(H X(t), R \right )
    \\
    Z(t)
    &\defeq
    f(H_0\mathbb{E}[X(t)]).
    \label{eq:measurement_model_II}
\end{align}
We will employ this measurement model in this paper.
The dynamic and measurement model together are called a \emph{probabilistic state space model} (SSM), which Gaussian ODE filtering uses to infer $x$ as detailed in \Cref{alg:godef}.
\begin{algorithm}[tb]
   \caption{Gaussian ODE Filtering}
   \label{alg:godef}
\begin{algorithmic}
   \STATE {\bfseries Input:} IVP$(x_i, m, T)$, step size $h>0$
   \STATE Initialize, $t=0$, $H_0X(0)=x_0$ and $HX(0)=f(x_0)$
   \REPEAT
       \STATE {\bfseries predict} state $X$, $t \to t+h$, along \Cref{eq:dynamic_model}
       \STATE $t = t+h$
       \STATE {\bfseries update} $X(t)$ by \Cref{eq:measurement_model,eq:measurement_model_II}
   \UNTIL{$t+h > T$}
\end{algorithmic}
\end{algorithm}  

{\bf The classical Taylor SSM} \

In previous research, the most commonly recommended model uses an integrated Brownian motion as a dynamic model (prior).
It is defined by inserting the following matrices into \Cref{eq:dynamic_model}:
\begin{align} \label{def:A^IOUP}
  A(h)_{ij}
  &=
  \mathbb{I}_{i\leq j} \frac{h^{j-i}}{(j-i)!},
  \\
  \label{eq:Q_IOUP}
  Q(h)_{ij}
  &=
  \sigma^2 \frac{h^{2q+1-i-j}}{(2q+1-i-j)(q-i)!(q-j)!},
\end{align}
where $\sigma^2$ is the variance scale of the underlying Brownian motion; see \citet[Appendix A]{KerstingSullivanHennig2018}.
Here, the state vector 
\begin{align}   \label{eq:X_IOUP}
    \left [ x^{(0)}(t), \dots, x^{(q)}(t) \right ]
    \ \sim \
    X(t)
\end{align} 
models the first $q \in \mathbb N$ derivatives of $x(t)$.
Therefore, the mean prediction $X(t+h) = A(h)X(t)$ is a \emph{Taylor expansion} of the numerical estimates of these derivatives.
Consequently, since Runge--Kutta methods also use Taylor approximations of $x(t)$ \citep[Section II.2]{hairer87:_solvin_ordin_differ_equat_i}, the posterior mean is similar to Runge--Kutta methods with local convergence rates of $q+1$ and global convergence rates of $q$.
It is hence a very good method to solve generic ODEs; see \citet{schober2019} and \citet{KerstingSullivanHennig2018}.

\section{Fourier Models for ODEs}

In this paper, we are concerned with oscillating ODEs. 
Hence, let the ODE be such that its solution $x(t)$ is a periodic function.
Let us denote the period of $x$ by $p > 0$, and its angular velocity by $w_0 = 2 \pi / p$.
For such periodic functions, a $J$-th order \emph{Fourier series}, $J \in \mathbb N$, of $x$ is the standard approximation:
\begin{align}
    \label{eq:fourier_sum}
    x^J(t)
    &=
    x_0
    +
    \sum_{j=1}^J x_j(t)
    \
    \stackrel{J \to \infty}{\to}
    x(t),
\end{align}
almost everywhere, where $x_0 = a_0/2$ and
\begin{align}   \label{eq:x_j_def}
    x_j(t)
    =
    a_j \cos(w_0 j t) + b_j \sin(w_0 j t).
\end{align} 
The derivative of \Cref{eq:fourier_sum} is 
\begin{align}
    \dot{x}^J(t)
    &=
    y_0
    +
    \sum_{j=1}^J
    - w_0 j y_j(t)
    \
    \stackrel{J \to \infty}{\to}
    x(t),
    \label{eq:der_fourier_sum}
\end{align}
almost everywhere, where $y_0 = 0$ and
\begin{align}   \label{eq:y_j_def}
    y_j(t) = a_j \sin(w_0 j t) - b_j \cos(w_0 j t).
\end{align}
The exact \emph{Fourier coefficients} are given by the integrals
\begin{align}
    \label{eq:Fourier_coefficient_a}
    a_j
    &=
    \frac{2}{p} \int_0^p x(t) \cos(j w_0 t) \ \rd t,
    \\
    b_j
    &=
    \frac{2}{p} \int_0^p x(t) \sin(j x_0 t) \ \rd t,
    \label{eq:Fourier_coefficient_b}
\end{align}
which are, in general, intractable.
Thus, learning a periodic approximation of $x$ amounts to inferring the coefficients $(a_j, b_j)$.
To reproduce the model from \citet[Section 3.2]{SolinSarkka_Periodic_2014}, we will however run inference on the corresponding harmonic oscillators $(x_j(t), y_j(t))$ instead.
To this end, we first observe that, for each $j=0,\dots,J$, $[x_j(t), y_j(t)]$ satisfies the following differential equations
\begin{align}   \label{eq:x_j_y_j_ode}
    \frac{\rd}{\rd t} 
    \begin{bmatrix}
    x_j(t) 
    \\ 
    y_j(t)
    \end{bmatrix}  
    =
    \begin{bmatrix}
        0 & -j w_0
        \\
        j w_0 & 0
    \end{bmatrix}
    \begin{bmatrix}
        x_j(t)
        \\
        y_j(t)
    \end{bmatrix}.
\end{align}
Note that, if we set $[x_0(0), y_0(0)] = [a_0/2,0]$ and $[x_j(0), y_j(0)] = [a_j, -b_j]$, then the only solution of \Cref{eq:x_j_y_j_ode} is indeed $[x_j(t), y_j(t)]$ as defined in \Cref{eq:x_j_def,eq:y_j_def}.
Since we (in general) do not know the Fourier coefficients $(a_j,b_j)$, we model the initial values with a Gaussian probability distribution: $[x_j(0), y_j(0)] \sim \mathcal{N}(\bs{0}, q_j^2 \bs{I})$, for some $q_j^2 > 0$.
We then model these oscillators by a stochastic process $X(t)$, i.e. 
\begin{align}  \label{eq:Fourier_state}
    [x_0(t), y_0(t), x_1(t), y_1(t), \dots, x_J(t), y_J(t)]
    \sim
    X(t)
\end{align}
which (according to \Cref{eq:x_j_y_j_ode}) follows the SDE, \Cref{eq:SDE}, if and only if
\begin{align}
    F
    &=
    \operatorname{diag}(F_1, \dots, F_J),
    \qquad
    \text{with blocks } 
    \\
    F_j
    &=
    \begin{bmatrix}
        0 & -j w_0
        \\
        j w_0 & 0
    \end{bmatrix},
    \qquad 
    \text{ and }
    \\
    L
    &=
    \boldsymbol{0}
    \
    \in
    \R^{2(J+1) \times 2(J+1)},
\end{align}
and if the initial condition is $[X(0)_{2j}, X(0)_{2j+1}] \sim \mathcal{N}(\bs{0},q_j^2 \bs{I})$.
It is natural that there is no diffusion ($L=0$), as the Fourier coefficients (unlike Taylor coefficients) of a periodic signal $x(t)$ do not change in $t$.
Since we want the prior defined by this SDE to be a zero-mean Gaussian Process with the canonical periodic covariance function
\begin{align}
    k_p(t,t^{\prime})
    =
    \sigma^2 \exp \left ( - \frac{2 \sin^2 \left ( w_0 \frac{t - t^{\prime}}{2}  \right ) }{l^2}   \right ),
\end{align}
we have to set 
\begin{align}
    q_j^2
    =
    \frac{2 I_j(l^{-2})}{\exp(l^{-2})},
    \qquad
    \text{ for }
    j=1,\dots,J,
\end{align}
where $I_j(z)$ is the modified Bessel function of the first kind of order $j$; see \citet[Eq.~27]{SolinSarkka_Periodic_2014}.
\\
{\bf Dynamic model} \
The implied matrices for the dynamic model, \Cref{eq:dynamic_model}, are now given by
\begin{align}
    A
    &=
    \operatorname{diag}(A_0, \dots, A_J),
    \qquad
    \text{with blocks } 
    \\
    A_j 
    &=
    \begin{bmatrix}
        \cos(w_0 j t) & -\sin(w_0 j t)
        \\
        \sin(w_0 j t) & \cos(w_0 j t)
    \end{bmatrix}, 
    \qquad
    \text{ and }
    \\
    Q
    &=
    \bs{0}
    \ \in
    \R^{2(J+1) \times 2(J+1)}.
\end{align}

{\bf Measurement Model} \
This dynamic model is, like all dynamic models, combinable with all measurement models. 
For ODEs, we need, by \Cref{eq:measurement_model}, a model $H_0$ that extracts $x(t)$ and a model $H$ that extracts the derivative from the state $X(t)$ of \Cref{eq:Fourier_state}.
By \Cref{eq:fourier_sum,eq:der_fourier_sum}, this is satisfied by
\begin{align}
    \label{eq:Fourier_H0}
    &H_0
    =
    \left [
        1 , 0 , 1 , 0 , \dots , 1 , 0
    \right ] \ \in \R^{1 \times 2(J+1)},
    \quad
    \text{and}
    \\
    H
    &=
    \begin{bmatrix}
        0 , 0 , 0 , -1w_0 , 0 , -2w_0 , \dots , 0 , -Jw_0 
    \end{bmatrix}
    \in \R^{1 \times 2(J+1)}
    .
    \notag
\end{align}

\subsection{Discussion of the Fourier model} 

As their coefficients do not depend on a support point, Fourier models are (unlike Taylor models) global expansions.
Hence, they are best at extrapolating globally, while Taylor methods excel at extrapolating locally.  
As ODE methods are usually designed for a small step size $h > 0$, the Taylor approximation is the standard approximation in ODE solvers, such as Runge--Kutta methods.
Hence, we expect the Fourier SSM to be more useful for global extrapolation with larger step sizes.
Accordingly, we suggest a hybrid ODE solver which combines the Taylor and the Fourier state space model in the next section.
This model could be used to extrapolate from a certain time, after learning the Fourier coefficients with data from the Taylor model.

\section{The Hybrid Taylor-Fourier Model}

As Taylor approximations excel at local approximations and Fourier approximations at global approximations of periodic signals, we combine both to the hybrid Taylor-Fourier model. 
The idea is that one can learn the Fourier coefficients $(a_j,b_j)$ from \Cref{eq:Fourier_coefficient_a,eq:Fourier_coefficient_b} with data from the Taylor SSM and then extrapolate with the Fourier SSM using the learned coefficients.
\hans{Maybe Gauss--Hermite quadrature can help here, as in [Mutny, Krause, 2018].}
Let us denote the Taylor SSM by $(A^{\text{Tay}}, Q^{\text{Tay}}, H^{\text{Tay}})$ and the Fourier SSM  $(A^{\text{Four}}, Q^{\text{Four}}, H^{\text{Four}})$. 
The hybrid Filter works now as follows:
It splits the time domain $[0,T]$ of the ODE into two parts $[0,T_p]$ and $[T_p,T]$ for some \emph{prediction time point} $T_p \in (0,T)$. 
On the first interval $[0,T_p]$, we solve the ODE with the classical Gaussian ODE filter with the Taylor SSM, and we, simultaneously, train the Fourier SSM with the data from the Taylor model.
On the second interval $[T_p,T]$, we just predict along the Fourier dynamical model defined by $(A^{\text{Four}}, Q^{\text{Four}})$.
\\
As the computation on the time interval $[T_p,T]$ does not require additional evaluations of $f$ and is therefore almost free, we hope that that this model turns out useful to reduce the computational time of solving periodic ODEs.
In the next section we present some experiments which, while not practical yet, highlight that this hybrid model in principle works.

\section{Experiments}

We try a Gaussian ODE filter with hybrid Taylor-Fourier SSM on two standard oscillating ODEs: the Van der Pol oscillator
\begin{align}   \label{eq:vdp}
    \begin{cases}
    \dot{x}_1(t) 
    &=
    \mu ( x_1(t) - \frac 13 x_1(t)^3 - x_2(t) ), 
    \\
    \dot{x}_2(t) 
    &=
    \frac{1}{\mu} x_1(t),
    \end{cases}
\end{align}
with $\mu = 5$, $x(0) = [1, -1]$, $T=50$, and the FitzHugh--Nagumo model
\begin{align}   \label{eq:fhn}
    \begin{cases}
        \dot{x}_1(t)
        &=
        x_1(t) - \frac{x_1(t)^3}{3} - x_2(t) + I,
        \\
        \dot{x}_2(t)
        &=
        \frac{1}{\tau} 
        \left ( x_1(t) + a - x_2(t) \right ),
    \end{cases}
\end{align} 
with parameters $(I,a,b,\tau) = (0.5,0.7,0.8,10.0)$, $x(0) = [1., 0.1]$ and $T=50$.

\subsection{Experimental Set-Up}

We set the parameters of the Taylor model as follows: $q=1$ and $\sigma^2 = 1$.
Moreover, we choose the following parameters of the Fourier model: $l=3$, $w_0 = 1$, $\sigma^2 = 1$, $J=3$  $p=2\pi$ and $R=0$.
Note that these parameters are not fine-tuned.
We define the prediction time $T_p$ for both ODEs to be $T_p = \frac{3}{4}T = 37.5$.

\begin{figure}[t]
    \setlength{\figwidth}{.9\columnwidth}
    \setlength{\figheight}{.5\figwidth}
    \centering\scriptsize
    \input{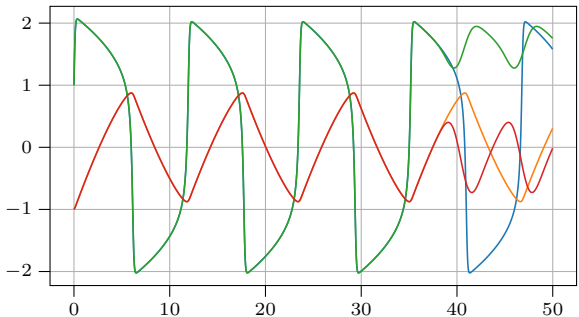}  
    \caption{Hybrid Filter on Van der Pol ODE, \Cref{eq:vdp}. Blue and yellow line are true curves $x_1$ and $x_2$. Red and green line are hybrid filter mean with prediction from $T_p = 37.5$. \label{ODE_vdp}
    }
\end{figure}

\begin{figure}[t]
    \setlength{\figwidth}{.9\columnwidth}
    \setlength{\figheight}{.5\figwidth}
    \centering\scriptsize
    \input{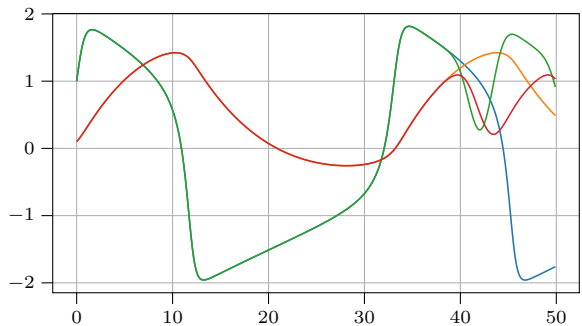}  
    \caption{Hybrid Filter on FitzHugh--Nagumo ODE, \Cref{eq:fhn}. Blue and yellow line are true curves $x_1$ and $x_2$. Red and green line are hybrid filter mean with prediction from $T_p = 37.5$. \label{ODE_fhn}
    }
\end{figure}

\subsection{Results}

The plots in \Cref{ODE_vdp,ODE_fhn} show that the hybrid ODE filter works in principle.
It picks up some structure from the trajectories on $[0,T_p]$ and can extrapolate forward by a sum of harmonic oscillators.
The quality of the extrapolation is, however, not good enough yet.
We suspect that this is due to our ad-hoc choice of parameters. 
Since our state space model is by \citet{SolinSarkka_Periodic_2014} an approximation of GP regression with periodic kernel and derivative observations, it should be possible to make the extrapolation as accurate as periodic GP regression (at least for $J \to \infty$).
\\
In particular, it should be possible to choose the angular velocity $w_0$ in a more principled way and thereby match the period of the oscillator more precisely.
Our choice of $w_0 = 1$ is particularly off in \Cref{ODE_fhn}.
We also believe that a more accurate extrapolation could be achieved if a larger $J$ is chosen. 
This will probably only work well once we have found a suitable way to choose $w_0$.
Moreover, future research should examine which $J+1$ data points from the Taylor model should be used for the Fourier model---which is an active learning task with Gaussian processes \citep{Seo_ActiveLearning_2000}.

\section{Conclusion}

We examined how Fourier state space models can be employed in Gaussian ODE filtering, to solve oscillating ODEs. 
To this end, we developed a novel Fourier sate space model that is applicable to ODEs.
We reasoned that it might outperform Taylor methods on global extrapolation tasks.
Since Fourier expansions are not locally accurate enough to serve as a practical ODE solver on its own, we have developed the hybrid ODE Solver which combines Taylor and Fourier expansions.
It first uses a Taylor SSM to compute up to a certain time $T_p$ while training a Fourier SSM `on the fly', and then uses the so-trained Fourier SSM to predict forward.
We demonstrated that, in principle, this can work---even if we are not yet satisfied with the quality of the Fourier prediction.
\\
Future research should examine how the Fourier coefficients can be learned better---e.g.~by finding ways to choose the Fourier parameters $(w_0,J)$ better or to employ smart active learning \citep{Seo_ActiveLearning_2000} for the Fourier coefficients $(a_j,b_j)$.
Since the desired Fourier coefficients are integrals, maybe ideas from Bayesian quadrature \citep{Briol2015probint} can be borrowed for this purpose; see \Cref{eq:Fourier_coefficient_a,eq:Fourier_coefficient_b}.
We hope that this might pave the way to almost cost-free predictions of oscillating systems which could come in useful in settings where ODEs have to be solved over a long time horizon with very limited budget or where a (reinforcement learning) system has to make sudden decisions in the context of ODE dynamics \citep{Deisenroth2011c}.
\\
If so, then exciting new ideas unknown to classical numerical analysis---such as quasi-periodic extrapolations \citep[Section 3.5]{SolinSarkka_Periodic_2014}---could be introduced to ODE solvers.
Such a development would also benefit probabilistic numerical methods for boundary value problems \citep{JohnSchober_GOODE_2019}, PDEs \citep{OatesCockayne_PN_PDEs_2019}, and ODE inverse problems \citep{KerstingKraemer_godef_inverse_2020}.
\\
In machine learning, such advances could provide better uncertainty quantification (of the numerical error) for continuous normalizing flows with ODEs, see \citep[Section 4]{ChenDuvenaud18}---where a free-form ODE, potentially an oscillator \citep{grathwohl2018scalable}, has to be numerically solved to approximate the transformed state and the numerical error is, to date, not accounted for. 

\bibliography{bibfile}

\begin{thebibliography}{18}
\providecommand{\natexlab}[1]{#1}
\providecommand{\url}[1]{\texttt{#1}}
\expandafter\ifx\csname urlstyle\endcsname\relax
  \providecommand{\doi}[1]{doi: #1}\else
  \providecommand{\doi}{doi: \begingroup \urlstyle{rm}\Url}\fi

\bibitem[Briol et~al.(2019)Briol, Oates, Girolami, Osborne, and
  Sejdinovic]{Briol2015probint}
Briol, F.-X., Oates, C.~J., Girolami, M., Osborne, M.~A., and Sejdinovic, D.
\newblock Probabilistic integration: A role for statisticians in numerical
  analysis? (with discussion and rejoinder).
\newblock \emph{Statistical Sciences}, 34\penalty0 (1):\penalty0 1--22
  (Rejoinder on p38--42), 2019.

\bibitem[Chen et~al.(2018)Chen, Rubanova, Bettencourt, and
  Duvenaud]{ChenDuvenaud18}
Chen, R., Rubanova, Y., Bettencourt, J., and Duvenaud, D.
\newblock Neural ordinary differential equations.
\newblock In \emph{Advances in Neural Information Processing Systems
  (NeurIPS)}, 2018.

\bibitem[Deisenroth \& Rasmussen(2011)Deisenroth and
  Rasmussen]{Deisenroth2011c}
Deisenroth, M. and Rasmussen, C.
\newblock {{PILCO}: A Model-Based and Data-Efficient Approach to Policy
  Search}.
\newblock In \emph{{International Conference on Machine Learning (ICML)}},
  2011.

\bibitem[Grathwohl et~al.(2019)Grathwohl, Chen, Bettencourt, and
  Duvenaud]{grathwohl2018scalable}
Grathwohl, W., Chen, R. T.~Q., Bettencourt, J., and Duvenaud, D.
\newblock Scalable reversible generative models with free-form continuous
  dynamics.
\newblock In \emph{International Conference on Learning Representations}, 2019.

\bibitem[Hairer et~al.(1987)Hairer, N{\o}rsett, and
  Wanner]{hairer87:_solvin_ordin_differ_equat_i}
Hairer, E., N{\o}rsett, S., and Wanner, G.
\newblock \emph{{Solving Ordinary Differential Equations I -- Nonstiff
  Problems}}.
\newblock Springer, 1987.

\bibitem[Hennig et~al.(2015)Hennig, Osborne, and Girolami]{HenOsbGirRSPA2015}
Hennig, P., Osborne, M.~A., and Girolami, M.
\newblock Probabilistic numerics and uncertainty in computations.
\newblock \emph{Proc. Roy. Soc. London A}, 471\penalty0 (2179):\penalty0
  20150142, 2015.

\bibitem[John et~al.(2019)John, Heuveline, and Schober]{JohnSchober_GOODE_2019}
John, D., Heuveline, V., and Schober, M.
\newblock {GOODE}: A {G}aussian off-the-shelf ordinary differential equation
  solver.
\newblock In \emph{International Conference on Machine Learning (ICML)}, 2019.

\bibitem[Kersting et~al.(2020{\natexlab{a}})Kersting, Kr\"amer, Schiegg,
  Daniel, Tiemann, and Hennig]{KerstingKraemer_godef_inverse_2020}
Kersting, H., Kr\"amer, N., Schiegg, M., Daniel, C., Tiemann, M., and Hennig,
  P.
\newblock Differentiable likelihoods for fast inversion of `likelihood-free'
  dynamical systems.
\newblock In \emph{International Conference on Machine Learning (ICML)},
  2020{\natexlab{a}}.

\bibitem[Kersting et~al.(2020{\natexlab{b}})Kersting, Sullivan, and
  Hennig]{KerstingSullivanHennig2018}
Kersting, H., Sullivan, T.~J., and Hennig, P.
\newblock Convergence rates of {Gaussian} {ODE} filters.
\newblock \emph{{arXiv:1807.09737v3 [math.NA]}}, 2020{\natexlab{b}}.

\bibitem[Oates et~al.(2019)Oates, Cockayne, Aykroyd, and
  Girolami]{OatesCockayne_PN_PDEs_2019}
Oates, C., Cockayne, J., Aykroyd, R., and Girolami, M.
\newblock {Bayesian} probabilistic numerical methods in time-dependent state
  estimation for industrial hydrocyclone equipment.
\newblock \emph{Journal of the American Statistical Association}, 114\penalty0
  (528):\penalty0 1518--1531, 2019.

\bibitem[Oates \& Sullivan(2019)Oates and Sullivan]{OatesSullivan2019}
Oates, C.~J. and Sullivan, T.~J.
\newblock A modern retrospective on probabilistic numerics.
\newblock \emph{Stat. Comput.}, 29\penalty0 (6):\penalty0 1335--1351, 2019.

\bibitem[Rackauckas et~al.(2018)Rackauckas, Ma, Dixit, Guo, Innes, Revels,
  Nyberg, and Ivaturi]{Rackauckas2018SensitivityAnalysisInML}
Rackauckas, C., Ma, Y., Dixit, V., Guo, X., Innes, M., Revels, J., Nyberg, J.,
  and Ivaturi, V.
\newblock A comparison of automatic differentiation and continuous sensitivity
  analysis for derivatives of differential equation solutions.
\newblock \emph{{arXiv:1812.01892 [math.NA]}}, 2018.

\bibitem[Rezende \& Mohamed(2015)Rezende and Mohamed]{RezendeMohamed_2015}
Rezende, D. and Mohamed, S.
\newblock Variational inference with normalizing flows.
\newblock In \emph{International Conference on Machine Learning (ICML)}, 2015.

\bibitem[Schober et~al.(2019)Schober, S{\"a}rkk{\"a}, and Hennig]{schober2019}
Schober, M., S{\"a}rkk{\"a}, S., and Hennig, P.
\newblock A probabilistic model for the numerical solution of initial value
  problems.
\newblock \emph{Stat. Comput.}, 29\penalty0 (1):\penalty0 99--122, 2019.

\bibitem[Seo et~al.(2000)Seo, Wallat, Graepel, and
  Obermayer]{Seo_ActiveLearning_2000}
Seo, S., Wallat, M., Graepel, T., and Obermayer, K.
\newblock {Gaussian} process regression: {Active} data selection and test point
  rejection.
\newblock \emph{Mustererkennung}, 2000.

\bibitem[Solin \& S{\"a}rkk{\"a}(2014)Solin and
  S{\"a}rkk{\"a}]{SolinSarkka_Periodic_2014}
Solin, A. and S{\"a}rkk{\"a}, S.
\newblock Explicit link between periodic covariance functions and state space
  models.
\newblock In \emph{Proceedings of the Seventeenth International Conference on
  Artificial Intelligence and Statistics}, volume~33 of \emph{Proceedings of
  Machine Learning Research}, pp.\  904--912. PMLR, 2014.

\bibitem[Tronarp et~al.(2019)Tronarp, Kersting, S{\"a}rkk{\"a}, and
  Hennig]{TronarpKSH2019}
Tronarp, F., Kersting, H., S{\"a}rkk{\"a}, S., and Hennig, P.
\newblock Probabilistic solutions to ordinary differential equations as
  nonlinear {Bayesian} filtering: a new perspective.
\newblock \emph{Stat. Comput.}, 29\penalty0 (6):\penalty0 1297--1315, 2019.

\bibitem[Tronarp et~al.(2020)Tronarp, Kersting, S{\"a}rkk{\"a}, and
  Hennig]{TronarpSarkkaHennig_BayesianODESolvers_2020}
Tronarp, F., Kersting, H., S{\"a}rkk{\"a}, S., and Hennig, P.
\newblock Bayesian ode solvers: The maximum a posteriori estimate.
\newblock \emph{{arXiv:2004.00623 [math.NA]}}, 2020.

\end{thebibliography}
\bibliographystyle{icml2020}

\end{document}